\documentclass{article} 
\usepackage{aaai25}  
\usepackage{times}  
\usepackage{helvet}  
\usepackage{courier}  
\usepackage[hyphens]{url}  
\usepackage{graphicx} 
\usepackage{natbib}  
\usepackage{caption} 
\usepackage{algorithm}
\usepackage{algorithmic}
\usepackage{multicol}
\usepackage{multirow}
\usepackage{amssymb}
\usepackage{amsmath}
\usepackage{booktabs}

\usepackage{newfloat}
\usepackage{listings}
\DeclareCaptionStyle{ruled}{labelfont=normalfont,labelsep=colon,strut=off} 
\floatstyle{ruled}
\newfloat{listing}{tb}{lst}{}
\floatname{listing}{Listing}
\newcommand*\samethanks[1][\value{footnote}]{\footnotemark[#1]}

\title{RealisID: Scale-Robust and Fine-Controllable Identity Customization \\via Local and Global Complementation}

\author {
    Zhaoyang Sun\textsuperscript{\rm 1,2}\thanks{Equal contribution. Work done during internship of Zhaoyang Sun at DAMO Academy, Alibaba Group} ,
    Fei Du\textsuperscript{\rm 2,3}\samethanks,
    Weihua Chen\textsuperscript{\rm 2,3},
	Fan Wang\textsuperscript{\rm 2,3}\\
	Yaxiong Chen\textsuperscript{\rm 1},
	Yi Rong\textsuperscript{\rm 1}\thanks{Corresponding author},
	Shengwu Xiong\textsuperscript{\rm4,5}\samethanks
}
\affiliations {
    \textsuperscript{\rm 1} Wuhan University of Technology \\
    \textsuperscript{\rm 2}DAMO Academy, Alibaba Group \\ 
	\textsuperscript{\rm 3}Hupan Laboratory  \\
	\textsuperscript{\rm 4}Shanghai AI Laboratory\\
	\textsuperscript{\rm 5}Interdisciplinary Artificial Intelligence Research Institute, Wuhan College, \\
	
}

\usepackage{bibentry}

\begin{document}
\maketitle

\begin{abstract}
	Recently, the success of text-to-image synthesis has greatly advanced the development of identity customization techniques, whose main goal is to produce realistic identity-specific photographs based on text prompts and reference face images. However, it is difficult for existing identity customization methods to simultaneously meet the various requirements of different real-world applications, including the identity fidelity of small face, the control of face location, pose and expression, as well as the customization of multiple persons. To this end, we propose a scale-robust and fine-controllable method, namely RealisID, which learns different control capabilities through the cooperation between a pair of local and global branches. Specifically, by using cropping and up-sampling operations to filter out face-irrelevant information, the local branch concentrates the fine control of facial details and the scale-robust identity fidelity within the face region. Meanwhile, the global branch manages the overall harmony of the entire image. It also controls the face location by taking the location guidance as input. As a result, RealisID can benefit from the complementarity of these two branches. Finally, by implementing our branches with two different variants of ControlNet, our method can be easily extended to handle multi-person customization, even only trained on single-person datasets. Extensive experiments and ablation studies indicate the effectiveness of RealisID and verify its ability in fulfilling all the requirements mentioned above.
\end{abstract}

\section{Introduction}
	The tremendous success of text-to-image synthesis techniques \cite{Imagen,GLIDE,DALL_E,StableIdentity} have spawned and driven a variety of customized generation tasks \cite{PaintByExample,StableVITON,MMDiff}. In this paper, we study one of the most prominent of these tasks, namely identity (ID) customization. It aims to adapt text-to-image synthesis models to generate new images, which match the identities depicted in the given reference face images and follow the controls prompted by the input text.
	
	Most early researches on ID customization typically fine-tune some specific parameters on a set of images containing the same ID so that the information associated with this ID can be integrated into the model. Although fine-tuning-based methods \cite{LoRA,DreamBooth,TextualInversion} have achieved commendable results, their computational costs are often significant. Even with advanced GPUs, fine-tuning the model parameters for every single ID can takes several minutes, which may make these methods infeasible in practical applications. To alleviate this issue, a series of recent works \cite{Fastcomposer,InstantID,FlashFace,PuLID} attempt to leverage ID-related prior knowledge learned from other large-scale face datasets to support fast inference without the need for fine-tuning. Generally, these approaches merge the ID features encoded from CLIP \cite{CLIP} or pre-trained face recognition model into the text embeddings or generative models by introducing trainable ID adapters \cite{IP_Adapter,T2I_Adapter,ControlNet}. 
	
\begin{figure*}[t]
		\centering
		\includegraphics[width=0.9\textwidth]{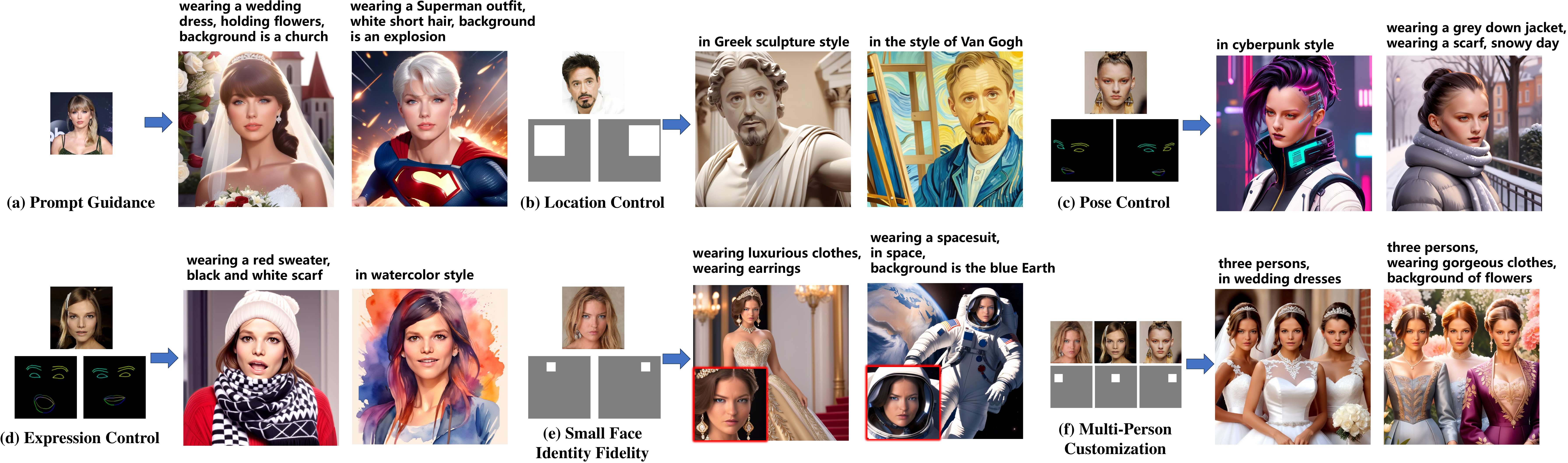}
		\captionof{figure}{Our RealisID can flexibly and finely control the face location, pose and expression factors of the generated facial images. It is also able to keep high identity fidelity for small faces and easily generalizes to multi-person customization.}
		\label{fig:requirements}
	\end{figure*}

	Despite the effectiveness and efficiency, all existing methods fail to simultaneously satisfy the various requirements (as illustrated in Fig.~\ref{fig:requirements}) of different real-world applications: (1) \textbf{Identity Fidelity of Small Face.} The faces in the generated images may need to be of different sizes. While most previous methods are effective in generating large faces, they often struggle with maintaining the identity details for small ones. (2) \textbf{Flexible and Fine Control.} In addition to identity information, we may also wish to finely adjust certain factors in the target images, such as face location, pose and expression. Since these factors cannot be precisely described through text, existing approaches that rely solely on text prompts will face challenges in appropriately controlling them. (3) \textbf{Multi-person Customization.} Most current methods are limited to customization for a single individual and lack the flexibility required for practical applications involving multiple persons. In addition, the scarcity of multi-person datasets available for model training further exacerbates the difficulty of meeting this requirement. 
	
	To this end, in this paper we propose a novel unified identity customization framework, namely RealisID, which is scale-robust and fine-controllable, thus can fulfill all the requirements mentioned above. Similar to the recently proposed state-of-the-art methods \cite{InstantID, PuLID, PhotoMaker}, our RealisID framework also attempts to inject additional condition signals into a pre-trained generative model (e.g., Stable Diffusion \cite{LDM, SDXL}). But different from all existing works, RealisID performs the condition information injection both locally and globally through two complementary branches. Specifically, we observe that the face-irrelevant contents in the reference image will prevent its latent space embedding from aligning well with the face-focusing control conditions (e.g., pose and expression). This may lead to difficulties in achieving the fine control of facial details and the identity fidelity for small faces. To address this problem, our local branch crops the face region from the latent embedding of the whole image, and up-samples it to the same spatial size. By taking this cropped embedding as input, the local branch will focus on injecting facial details information. And also, the up-sampling operation narrows the face scale differences across different reference images, making our local branch to be scale-robust. In addition, we also design a global branch to integrate face location information and manage the overall harmony of the entire image. The complementarity of these two branches allows RealisID to generate high-quality facial images that can be flexibly manipulated. Furthermore, by implementing our two branches with the ControlNet \cite{ControlNet}, RealisID can be easily extend to handle multi-person customization, even without training on multi-person data.
	Our main contributions are summarized as follows:

	\begin{itemize}
	  \item We introduce RealisID, which achieves scale-robust and fine-controllable ID customization through the collaborative efforts of two complementary branches.
	  \item To the best of our knowledge, by equipping the newly proposed local branch, our RealisID is the first attempt for scale-robust ID customization, which can effectively maintain the identity fidelity for small faces.
	  \item Extensive comparisons indicate that RealisID outperforms five recent state-of-the-art ID customization methods, especially in the small face scenario. And ablation studies validate the ability of RealisID in facial factors fine control and multi-person customization.
	\end{itemize}

	\begin{figure*}[t]
	  \centering
	  \includegraphics[width=0.90\linewidth]{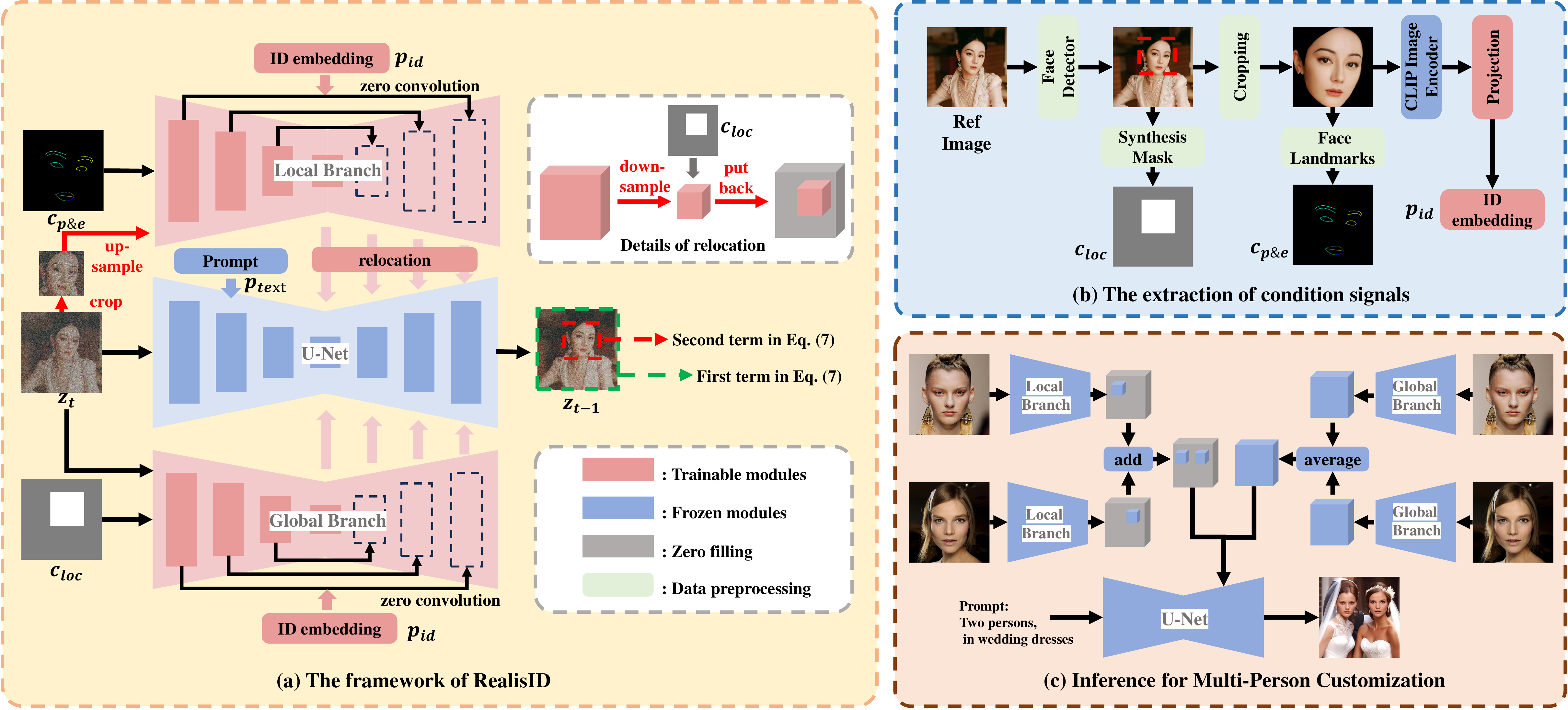}
	  \caption{(a) The overall architecture of our RealisID framework, which constructs a pair of local and global branches to inject additional condition information into the U-Net denoiser of a pre-trained stable diffusion model. (b) The procedure of extracting different condition signals from the input reference images. (c) The inference strategy for handling multi-person customization.
	  }
	  \label{fig:method}
	\end{figure*}

\section{Related Works}
\subsubsection{Text-to-Image Synthesis}
Empowered by diffusion models, text-to-image synthesis techniques have achieved significant advances. The typical procedure involves encoding textual prompts into latent vectors using pre-trained text encoders such as CLIP \cite{CLIP} to steer the diffusion process. Imagen \cite{Imagen}, GLIDE \cite{GLIDE}, and DALL-E \cite{DALL_E}  perform denoising directly in the original pixel space. Stable Diffusion, a standout work in latent diffusion models \cite{LDM}, utilizes autoencoders for denoising in latent space to optimize computational efficiency while maintaining the ability to generate high-quality images. In continuation, SDXL \cite{SDXL} introduces a larger UNet, a refiner model, and a second text encoder to bolster image quality and text control. Owing to the remarkable achievements in image synthesis, text-to-image models have been widely used as the backbone for many generative tasks.

\subsubsection{Identity Customization} 
 ID customization is a challenging task in subject-driven image generation. Current research can be categorized into two groups based on the necessity of fine-tuning during inference. Fine-tuning methods \cite{LoRA,DreamBooth,TextualInversion,StableIdentity} involve adjusting parameters on multiple images with the same ID or injecting the ID condition into the model using inversion techniques. However, optimization-based approaches require individual training for each new role, leading to substantial computational expenses and constraining flexibility and practicality. In contrast, fine-tuning free  methods show promising results on zero-shot identity customization. These methods \cite{Fastcomposer,Caphuman,IDAdapter,PhotoMaker,PortraitBooth,InstantID,FlashFace,PuLID} initially encode reference facial features into one or more labels, then train additional adapters (such as IP-Adapter, ControlNet) and integrate identity conditioning into the pretrained text-to-image model. However, these methods may encounter poor identity fidelity in small faces, lack precise and adaptable facial control, or fail to scale to multi-person customization. To overcome these challenges, this paper introduces RealisID, which eliminates the need for fine-tuning during the inference stage and can be seamlessly extended to handle multi-person customization task \cite{InstantFamily}.

\section{Methodology}
\subsection{Preliminary}
	Since our RealisID framework is built based on Stable Diffusion and ControlNet models, we briefly review them first.
	
	\textbf{Stable Diffusion (SD)} is a large-scale text-to-image latent diffusion model trained on LAION \cite{LAION_5B} dataset. It consists of three modules: The encoder $\mathcal{E}$ and the decoder $\mathcal{D}$ transform the input images into latent space and back to pixel space, respectively, playing a crucial role in reducing the computational complexity. And the U-Net denoiser $\epsilon_{\theta}$ is trained to predict the applied diffusion noise in the latent space as follow:
	\begin{equation}
		\label{equ1}
		\mathcal{L}_{sd} =\mathbb{E}_{z_{t},p,\epsilon \sim  \mathcal{N} (0,1), t \sim \mathcal{U} (1,T)} [\parallel\epsilon_{\theta }(z_{t},p,t)-\epsilon \parallel_{2}],
	\end{equation}
	where $z_{t}$ is the noisy latent variable at the $t$-th timestep, $p$ is a prompt signal. The noise $\epsilon$ is sampled from the standard Gaussian distribution and $T$ denotes the maximum timestep. 
	
	\textbf{ControlNet} is designed to integrate additional control conditions into pre-trained generative models (e.g., Stable Diffusion). Specifically, it injects the following condition information into the intermediate representation produced by each decoding layer of the U-Net denoiser $\epsilon_{\theta}$:
	\begin{equation}
		\label{eq:controlnet}
		i_{t} = \mathcal{Z}(\mathcal{F}(z_{t}+\mathcal{Z}(c; \Theta_{z1}	),p,t; \Theta_{c}); \Theta_{z2}),
	\end{equation}
	where $c$ is the input condition. $\mathcal{Z}(\cdot; \Theta_{z1})$ and $\mathcal{Z}(\cdot; \Theta_{z2})$ indicate two different zero convolution layers with parameters $\Theta_{z1}$ and $\Theta_{z2}$, respectively. $\mathcal{F}(\cdot, \cdot, \cdot; \Theta_{c})$ consists of trainable copy blocks whose parameter is $\Theta_{c}$.
	
\subsection{The Proposed RealisID Framework}
	Our RealisID framework consists of two branches, whose structures are two different variants of ControlNet. As illustrated in Fig.~\ref{fig:method}(a), these two branches introduce distinct condition control information to the pre-trained Stable Diffusion U-Net denoiser, respectively. The local branch focuses on injecting facial details information within the local face region, such as head pose and facial expressions as well as person identity. In contrast, the global branch manages the overall harmony of the entire image, with the face location and the corresponding body and background layouts being manipulated through this branch. The cooperation of these two branches enables the U-Net denoiser to exploit both of their complementary information, thus allowing the generative process to be flexibly and finely controlled.
	
	\subsubsection{Condition Signals for Model Training}
	In order to train both branches in our RealisID framework and achieve the corresponding control capabilities mentioned above, we first need to obtain the relevant condition signals from the training samples. Specifically, as shown in Fig.~\ref{fig:method}(b), given a reference image and its associated description text, we extract the following three types of condition signals from them:
	
	\textbf{(1) ID and Text Embeddings.} For each reference image, we first apply a face detection model to generate a bounding box that locates the face region, which is then manually adjusted into a square based on its longer edge. After that, the input image is cropped according to the obtained bounding box, and a pre-trained face parsing model is used to zero-out its irrelevant background areas. Subsequently, this cropped face image is fed into the CLIP \cite{CLIP} image encoder to extract image features from the penultimate hidden layer. Finally, we obtain the ID embedding $p_{id}$ of the reference image by aligning these extracted CLIP features with the U-Net latent space through a projection layer. For the input text, we use the same operations as SDXL \cite{SDXL} to get the corresponding text embedding $p_{text}$. During the model training, $p_{id}$ acts as the prompt signal of both local and global branches in our framework, and $p_{text}$ is fed into the U-Net denoiser as the text prompt input.
	
	\textbf{(2) Pose-Expression Representation.} 
	Apart from identity, the head pose and facial expression are also essential elements of the human face. To achieve the control of these factors, inspired by \cite{Emoji, AniPortrait}, we take the facial landmarks of the above cropped face image as the pose-expression representation $c_{p\&e}$. It will be utilized as the condition input of our local branch.
	
	
	\textbf{(3) Location Guidance.} 
	Based on the squared bounding box, we can generate a mask as the face location guidance $c_{loc}$. This mask is a single-channel matrix that fills the region inside the facial bounding box with 1 and other areas with 0. In addition to face region, $c_{loc}$ also implicitly indicates the locations of human body and image background. For model training, $c_{loc}$ is input into the global branch as the condition, and is also used for relocation operations in the local branch.
	
	\subsubsection{Local Branch}
	\label{sec:local_branch}
	According to Eq.~(\ref{eq:controlnet}), with the ID embedding $p_{id}$ as the prompt signal, the pose-expression representation $c_{p\&e}$ as the condition, and the noisy latent variable $z_{t}$, the injection information of our local branch can be produced as follow:
	\begin{equation}
		\label{eq:local1}
		i_{t\_l} = \mathcal{Z}(\mathcal{F}(z_{t}+\mathcal{Z}(c_{p\&e}; \Theta_{z1\_l}),p_{id},t; \Theta_{c\_l}); \Theta_{z2\_l}).
	\end{equation}
	By observing Eq.~(\ref{eq:local1}), we can find that the face-irrelevant information in $z_{t}$ will significantly influence the effectiveness of $i_{t\_l}$. This problem can be even worse when the face region becomes relatively smaller in the whole reference image. In such case, the face-irrelevant information will dominate $z_{t}$, making it unable to align with the pose-expression condition information $\mathcal{Z}(c_{p\&e}; \Theta_{z1})$, thus decreasing the ability of $i_{t\_l}$ for fine control and identity preservation. To deal with this problem, we takes cues from previous studies \cite{SodMTGAN,BetterToFollow} on improving small object detection through super-resolution. Specifically, instead of using the entire $z_{t}$, we crop and bilinearly up-sample the face area from it, and then take the obtained latent embedding $\hat{z}_{t}$ as the input of our local branch network. In this way, $\hat{z}_{t}$ will mainly focus on facial details information and can be well aligned with $\mathcal{Z}(c_{p\&e}; \Theta_{z1})$. And moreover, since the face regions in different reference images are up-sampled to the same input size, their scale differences can be effectively reduced, \textbf{making our local branch to be robust to face scale.} However, cropping the face from $z_{t}$ will result in the loss of its location information. To address this, we further insert a relocation operation before the information injection, which can be formulated as:
	\begin{equation}
		\label{eq:local2}
		\hat{i}_{t\_l} = \mathcal{R}(i_{t\_l}, c_{loc}),
	\end{equation}
	\begin{equation}
		\label{eq:local3}
		i_{t\_l} = \mathcal{Z}(\mathcal{F}(\hat{z}_{t}+\mathcal{Z}(c_{p\&e}; \Theta_{z1\_l}),p_{id},t; \Theta_{c\_l}); \Theta_{z2\_l}).
	\end{equation}
	As shown in Fig.~\ref{fig:method}, the relocation operation $\mathcal{R}(i_{t\_l}, c_{loc})$ first down-samples $i_{t\_l}$ according to the relative size of the face region to the entire input image. Then based on the location guidance $c_{loc}$, it puts the down-sampled features back to the face position in a zero tensor of the same size as $i_{t\_l}$. Finally, the obtained $\hat{i}_{t\_l}$ is injected into the U-Net desnoier to achieve the flexible and fine control of facial details as well as the scale-robust identity fidelity.
	
	\subsubsection{Global Branch}
	\label{sec:global_branch}
	By taking the ID embedding $p_{id}$ as the prompt signal, the location guidance $c_{loc}$ as the condition, and the entire noisy latent variable $z_{t}$, our global branch generates the following injection information:
	\begin{equation}
		\label{eq:global1}
		i_{t\_g} = \mathcal{Z}(\mathcal{F}(z_{t}+\mathcal{Z}(c_{loc}; \Theta_{z1\_g}),p_{id},t; \Theta_{c\_g}); \Theta_{z2\_g}).
	\end{equation}
	Based on the face region, $c_{loc}$ can implicitly indicate the locations of human body and image background, therefore it can provide the overall layout information of the whole image for $i_{t\_g}$. As a result, by injecting $i_{t\_g}$ in the U-Net desnoier, we can accurately control the face location and obtain harmonious generation results.
	
	\subsubsection{Training Loss}
	During the training phase, we only update the parameters of our two branches and the projection layer for generating the ID embedding, while freezing those of the pre-trained Stable Diffusion model. To this end, we optimize the following objective function:
	\begin{align}
	\label{eq:total_loss}	
		\mathcal{L}&=\mathbb{E}_{z_{t},p_{text},\epsilon \sim \mathcal{N} (0,1), t\sim\mathcal{U}(1,T)} [\parallel\epsilon_{\theta }(z_{t},p_{text},t)-\epsilon \parallel_{2} \nonumber \\
		&+\lambda\parallel(\epsilon_{\theta}(z_{t},p_{text},t)-\epsilon) \odot c_{loc} \parallel_{2}].
	\end{align}
	Here, $\odot$ denotes the element-wise multiplication operation. The first term in Eq.~(\ref{eq:total_loss}) is a standard stable diffusion loss, and the second term only calculates the noise discrepancy within the face area indicated by $c_{loc}$, thus facilitating the generation of facial details. $\lambda>0$ is a positive hyperparameter that balances the importance of these two terms.

\subsubsection{Inference Strategy} 
	  Following \cite{FlashFace}, we employ the classifier-free guidance scheme \cite{CFG} for model inference. It combines different types of noise predicted under three conditions: without ID embedding nor text prompt $\epsilon_{none}$, with only text prompt $\epsilon_{t}$, and with both ID embedding and text prompt $\epsilon_{t\&i}$, as follow:
	  \begin{equation}
	  	\label{eq:CFG}
	  	\epsilon_{prd}=\epsilon_{none}+\lambda_{t} (\epsilon_{t}-\epsilon_{none})+\lambda_{i}(\epsilon_{t\&i}-\epsilon_{t}),
	  \end{equation}
	  where $\epsilon_{prd}$ denotes the final noise prediction for inference procedure. $\lambda_{t}>0$ and $\lambda_{i}>0$ serve as two weighting factors for text and image guidance, respectively.

\subsubsection{Inference for Multi-Person Customization} 
	For the customization of multiple persons, given the reference image and the target location guidance of each individual, we need to integrate the injection information for all these conditions that are produced by our two branches, respectively. As illustrated in Fig.~\ref{fig:method}(c), for the local branch, the injection information of different reference images are simply summed, since each of them only affects the local areas of the whole image, while those generated by the global branch are averaged. The integrated information is then injected into the U-Net denoiser for noise prediction.

	\begin{figure*}[t]
		\centering
		\includegraphics[width=0.75\linewidth]{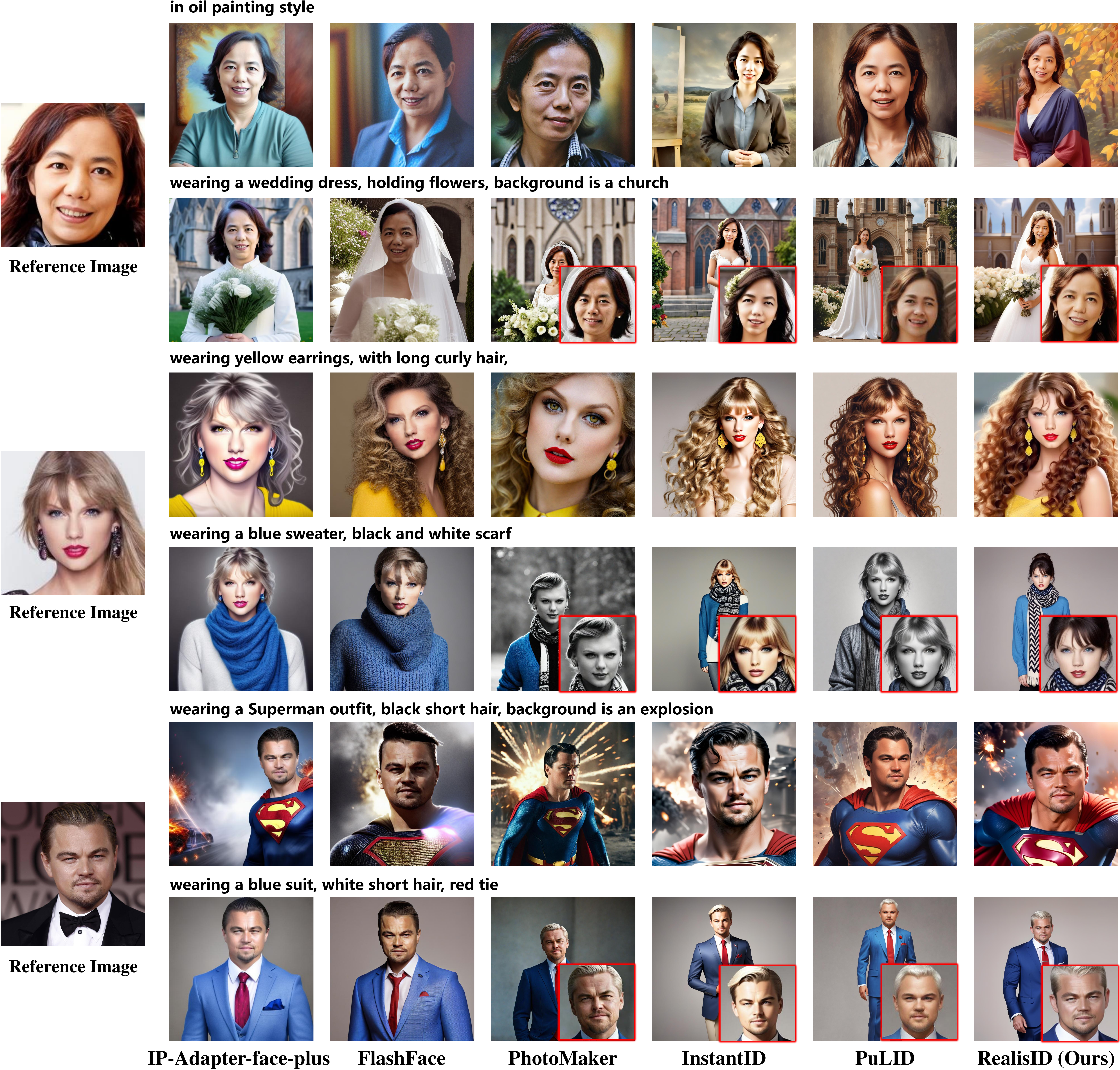}
		\caption{Qualitative comparison between different methods. The odd and even rows correspond to regular and small face scenarios, respectively. Regardless of face scales, our RealisID framework achieves the high fidelity of identity and facial details, thus generating visually appealing portrait images.}
		\label{fig:qualitative}
	\end{figure*}

\section{Experiments}

\subsection{Implementation Details}
	To improve the realism of generated results, we establish our framework based on the pre-trained SDXL-1.0 \cite{SDXL} text-to-image synthesis model. The trainable parameters in our RealisID model are learned from the publicly available CosmicMan dataset \cite{CosmicMan}, which comprises 2 million image-text pairs of single individuals. All the reference images are manually cropped and resized to 1024$\times$1024 pixels. To extract the condition signals from image samples, we use MTCNN \cite{MTCNN} for face detection, BiSeNet \cite{BiSeNet} for face parsing and MediaPipe \cite{MediaPipe} for facial landmark detection. During the training phase, we follow the learning strategy of IP-Adapter \cite{IP_Adapter} that randomly drops either the image prompt (i.e., ID embedding) or the text prompt or both of them with a probability of 0.05. The hyperparameter $\lambda$ in Eq.~(\ref{eq:total_loss}) is set to 1.0. The framework is optimized on 8 NVIDIA H20 GPUs, using the Adam optimizer with batch size of 16, learning rate of 1e-5 and weight decay of 1e-2. For inference, we adopt the same delayed subject conditioning technique as in \cite{Fastcomposer}. We set $\lambda_{t}=7.5$ and $\lambda_{i}=5.0$ in Eq.~(\ref{eq:CFG}), and use a 30-step DDIM \cite{DDIM} sampler to generate the target images.

	\begin{table*}[t]
		\centering
		\resizebox{0.8\linewidth}{!}{
			\begin{tabular}{l|c|c|c|c|c|c|c|c}
				\toprule 
				\multirow{2}{*}{Methods} & \multicolumn{4}{c|}{Regular Case}& \multicolumn{4}{c}{Small Face}  \\
				\cmidrule{2-5} \cmidrule{6-9} 
				& ASP $\uparrow$ & CLIP-T $\uparrow$ & FaceNet $\uparrow$  & CLIP-I $\uparrow$ & ASP $\uparrow$ & CLIP-T $\uparrow$ & FaceNet $\uparrow$  & CLIP-I $\uparrow$  \\
				\midrule
				IP-Adapter-face-plus &5.64 & 0.201 & 0.760 &0.714  & - &- &- &- \\
				FlashFace &5.46 & 0.212 & \textbf{0.809} &\textbf{0.754}  & - &- &- &- \\
				PhotoMaker &5.74 & 0.223 & 0.508 &0.651  & 5.78 &0.229 &0.516 & 0.658 \\
				InstantID &6.01 & 0.211 & 0.768 &0.706  & 5.72 &0.210 &\underline{0.693} &\underline{0.686} \\
				PuLID &\textbf{6.37} & \textbf{0.254} & 0.772 &0.697  & \underline{5.95} & \textbf{0.249} &0.497 &0.585 \\
				RealisID (Ours) &\underline{6.22} & \underline{0.234} & \underline{0.796} &\underline{0.739}  & \textbf{6.11} &\underline{0.236} &\textbf{0.767} &\textbf{0.701} \\
				\bottomrule
			\end{tabular}
		}
	    \caption{Quantitative comparison between different methods. Four metrics ASP, CLIP-T, FaceNet, and CLIP-I are calculated for both regular and small face cases. The best and second best results are highlighted in \textbf{bold} and \underline{underlined}, respectively.}
	    \label{table:quantitative}
	\end{table*}

\subsection{Baseline Methods}
	We compare our RealisID framework with five recently proposed state-of-the-art ID customization methods, including IP-Adapter-face-plus \cite{IP_Adapter}, FlashFace \cite{FlashFace}, PhotoMaker \cite{PhotoMaker}, InstantID \cite{InstantID}, and PuLID \cite{PuLID}. All these methods are implemented with their officially released code and pre-trained model parameters. Notably, our method keeps the same setting with PhotoMaker, InstantID and PuLID that utilizes SDXL-1.0 as the basic model, while IP-Adapter-face-plus and FlashFace are constructed based on SD-1.5. Following InstantID, when only a reference image is provided, the facial landmarks of the reference image are used as the pose-expression control condition input for our model.

\begin{table}[t]
	\centering
	\resizebox{0.8\linewidth}{!}{
		\begin{tabular}{l|c|c|c|c}
			\toprule
			\multirow{2}{*}{Methods} & \multicolumn{4}{c}{Small Face}  \\
			\cmidrule{2-5} 
			& ASP   & CLIP-T  & FaceNet & CLIP-I  \\
			\midrule
			w/o $B_{local}$ &6.07 & \textbf{0.243} & 0.681 &0.673   \\
			w/o $B_{global}$ &5.97 & 0.241 & 0.734 &0.689   \\
			Full Model &\textbf{6.11} & 0.236 & \textbf{0.767} & \textbf{0.701}  \\
			\bottomrule
		\end{tabular}
	}
	\caption{Effects of each individual branch in RealisID.}
	\label{table:ablation}
\end{table}

\subsection{Qualitative Comparisons}
	\subsubsection{Regular Case} In Fig.~\ref{fig:qualitative}, the odd rows display the quantitative results generated by different competing methods in a regular scenario, without involving particular conditions other than the input image and text. We can see that all these methods except PhotoMaker exhibit high identity fidelity. It is mainly because that PhotoMaker projects the identity features into the text embedding space that is contextually ambiguous, thus resulting in the loss of identity detail information. Moreover, the SDXL-1.0-based methods typically produce visually appealing and high-quality facial images due to the more powerful generation capabilities of SDXL-1.0. 
	
	\subsubsection{Images with Small Faces} 
	Ensuring the identity fidelity for faces of different sizes is a critical requirement for ID customization. The even rows in Fig.~\ref{fig:qualitative} present the quantitative comparisons of all competing approaches in generating target images with small faces. \emph{In this study, we define the ``small face'' as a human face that enclosed in a bounding box whose long side is less than 1/6 of the image edge.} For the methods that cannot control the face size, we introduce an additional text prompt ``a full-body people image'' to guide them in generating outcomes with small faces. From Fig.~\ref{fig:qualitative}, we observe that IP-Adapter-face-plus and FlashFace do not have the ability to synthesize small face images. PhotoMaker, InstantID and PuLID also experience a significant degradation in identity fidelity when compared to their results in the regular case. In contrast, our RealisID framework consistently maintains high identity fidelity for small faces. This is mainly attributed to the detail fine control ability and the scale robustness of our local branch, demonstrating the effectiveness of its scale alignment (i.e., face cropping and up-sampling) and feature relocation operations.

\subsection{Quantitative Comparisons}
\subsubsection{Evaluation Dataset} 
	Our evaluation data consists of 40 unseen identities obtained from another CelebA-HQ \cite{CelebA_HQ} dataset. For a comprehensive evaluation, we prepare 35 prompts that covering cover a variety of clothing, attributes, actions, backgrounds, and styles. Every method generates 2 images for each prompt of each identity, resulting in a total of 2800 synthesized images for evaluation. Please see the supplementary materials for more details.

\subsubsection{Evaluation Metrics} 
	Similar to the previous work \cite{PhotoMaker}, we use CLIP-T \cite{CLIP} to measure the prompt fidelity of the generated images and use ASP\footnote[1]{https://github.com/christophschuhmann/improved-aesthetic-predictor} metric to evaluate their aesthetic. We also employ FaceNet \cite{FaceNet} and CLIP-I \cite{TextualInversion} to assess the identity fidelity. Specifically, we calculate the identity similarity between the generated image and the reference image based on the face embeddings extracted by FaceNet from the face regions detected by MTCNN \cite{MTCNN}. For all these metrics, the higher value indicates the better generative performance.
	
	\begin{table}[t]
		\centering
		\resizebox{0.8\linewidth}{!}{
			\begin{tabular}{l|c|c|c|c}
				\toprule
				\multirow{2}{*}{Methods} & \multicolumn{4}{c}{Face Relative Size}  \\
				\cmidrule{2-5}
	      & 1/4   & 1/5  & 1/6 & 1/7 \\
				\midrule
				InstantID &0.765 &0.745 & 0.708&0.664   \\
	      RealisID (Ours) &\textbf{0.791} & \textbf{0.787}&\textbf{0.772}& \textbf{0.748} \\
				\bottomrule
			\end{tabular}
		}
		\caption{Identity fidelity (FaceNet) of different face sizes.}
		\label{table:scale}
	\end{table}

\subsubsection{Quantitative Results} 
	For the methods capable of controlling the face size, we randomly set the relative size of face regions (i.e., the ratio of face bounding box long side length to the whole image edge length) in the range of $[1/4, 1/2]$ for the regular scenario, and set those in the range of $[1/7, 1/6]$ for the small face scenario. Similar to our quantitative analysis, for the methods that cannot manipulate the face size, an additional text prompt of ``a full-body people image'' is introduced for small face image generation. The quantitative results are presented in Table~\ref{table:quantitative}. Since IP-Adapter-face-plus and FlashFace fail to generate small face images (see Fig.~\ref{fig:qualitative}), their corresponding evaluation metrics are not calculated. It can be seen that our RealisID achieves comparable results with other state-of-the-art methods in the regular case, which is consistently the second best method on all four metrics. Furthermore, in the small face scenario, RealisID showcases its superiority by achieving the best results on three out of four metrics. In particular, the significant improvements on FaceNet (0.767 vs. 0.693) and CLIP-I (0.701 vs. 0.686) indicate the effectiveness of our framework in maintaining high identity fidelity for small faces.

\subsection{Ablation Studies}
\subsubsection{Complementarity of Local and Global Branches} 
\begin{figure}[t]
	  \centering
	  \includegraphics[width=0.8\linewidth]{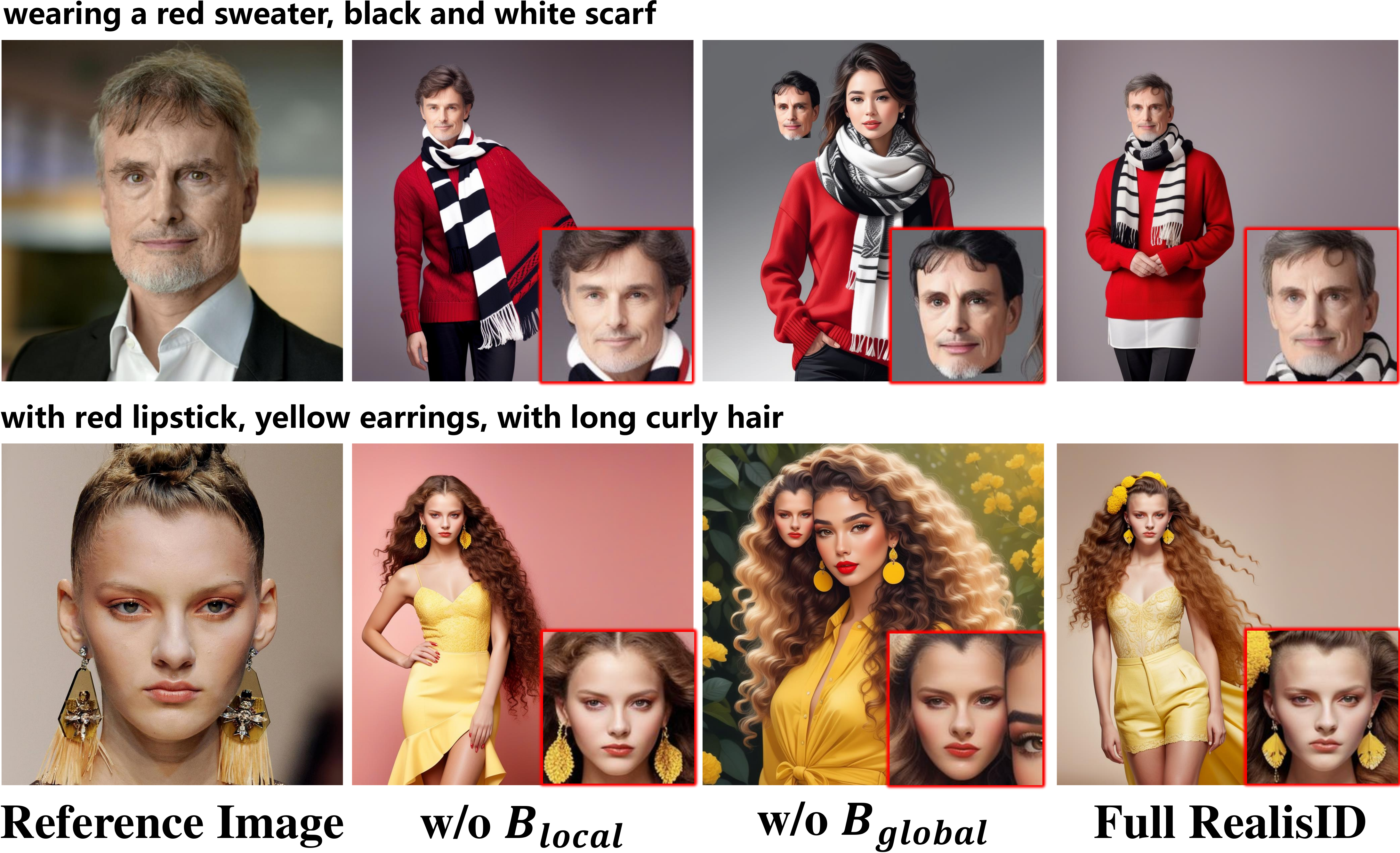}
	  \caption{Effects of our local and global branches.}
	  \label{fig:ablation}
	\end{figure}

	We construct two ablated models by separately removing the local and global branches from our full RealisID framework to investigate their effects on final generated results. Here, we use $B_{local}$ and $B_{global}$ to denote the two branches, respectively. We conduct this ablation study only in the small face scenario. As shown in Fig.~\ref{fig:ablation}, the removal of $B_{local}$ leads to distortions in identity and facial details, which is also reflected in the reduction of FaceNet and CLIP-I metrics in Table~\ref{table:ablation}. This suggests that $B_{local}$ plays an important role in preserving and controlling face-relevant details information. When $B_{global}$ is removed, the ablated model indeed synthesizes a face with higher identity fidelity (FaceNet: 0.734 and CLIP-I: 0.689) at the specified location. But this face is isolated from other image contents and does not blend well with the overall layout, thus achieving the lowest ASP value of 5.97. This demonstrates that $B_{global}$ manages the global harmony of the entire image. In contrast, benefiting from the complementarity and cooperation between $B_{local}$ and $B_{global}$, the full RealisID model produces more satisfactory and realistic results, both qualitatively and quantitatively.

	\begin{figure}[t]
	  \centering
	  \includegraphics[width=0.85\linewidth]{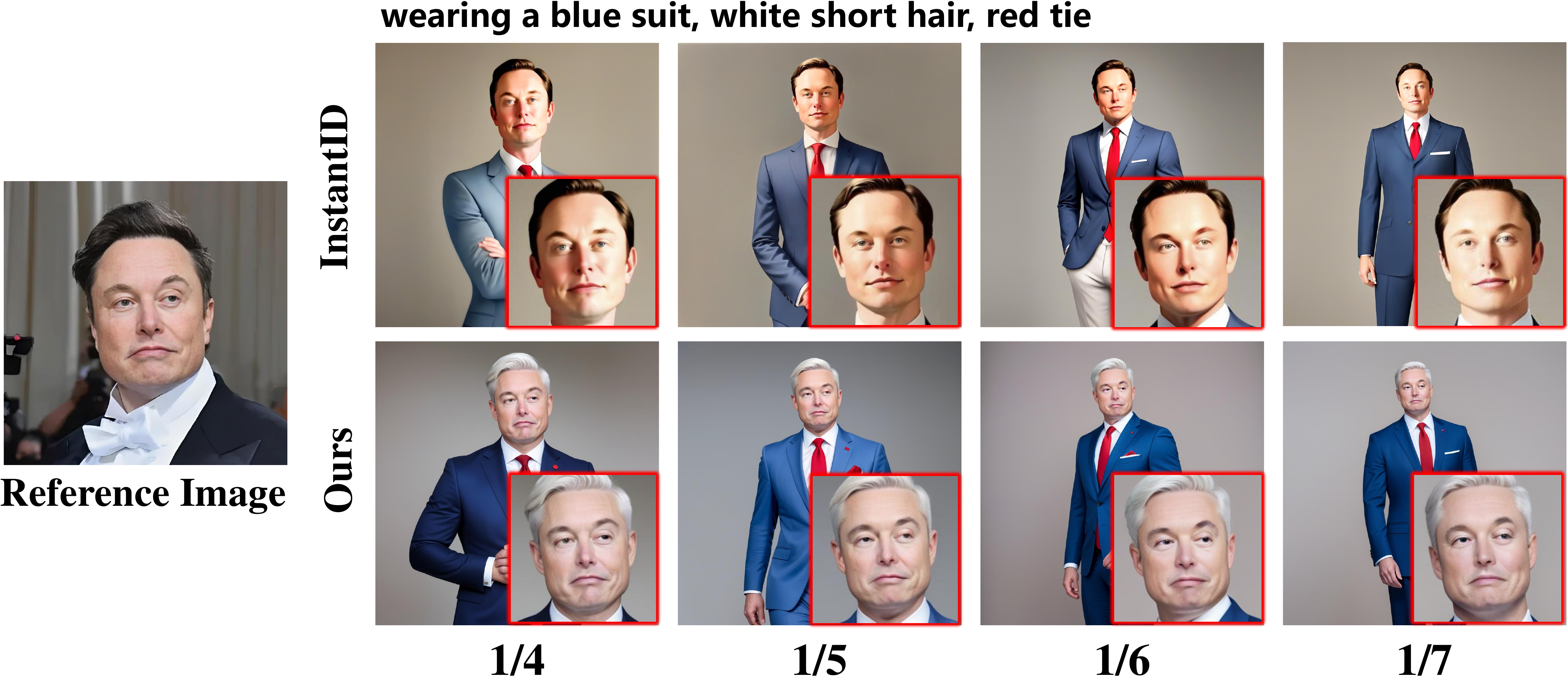}
	  \caption{Ablation study on scale robustness. Our method achieves high identity fidelity across different face sizes.
	  }
	  \label{fig:scale}
	\end{figure}
	
	\begin{figure}[t]
	  \centering
	  \includegraphics[width=0.85\linewidth]{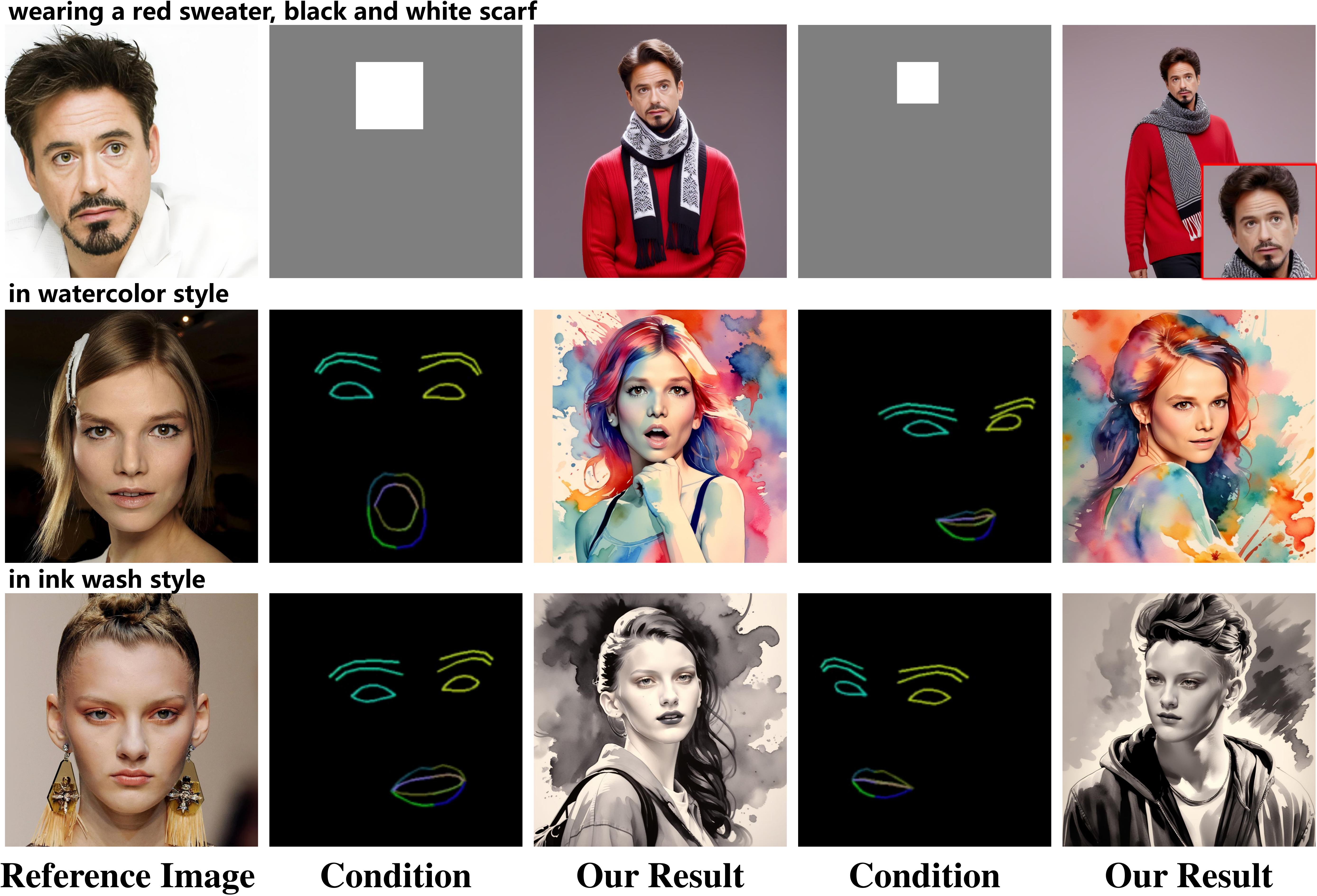}
	  \caption{Fine control of location, size, pose and expression.
	  }
	  \label{fig:pose_exp}
	\end{figure}
	
	\begin{figure}[t]
	  \centering
	  \includegraphics[width=0.8\linewidth]{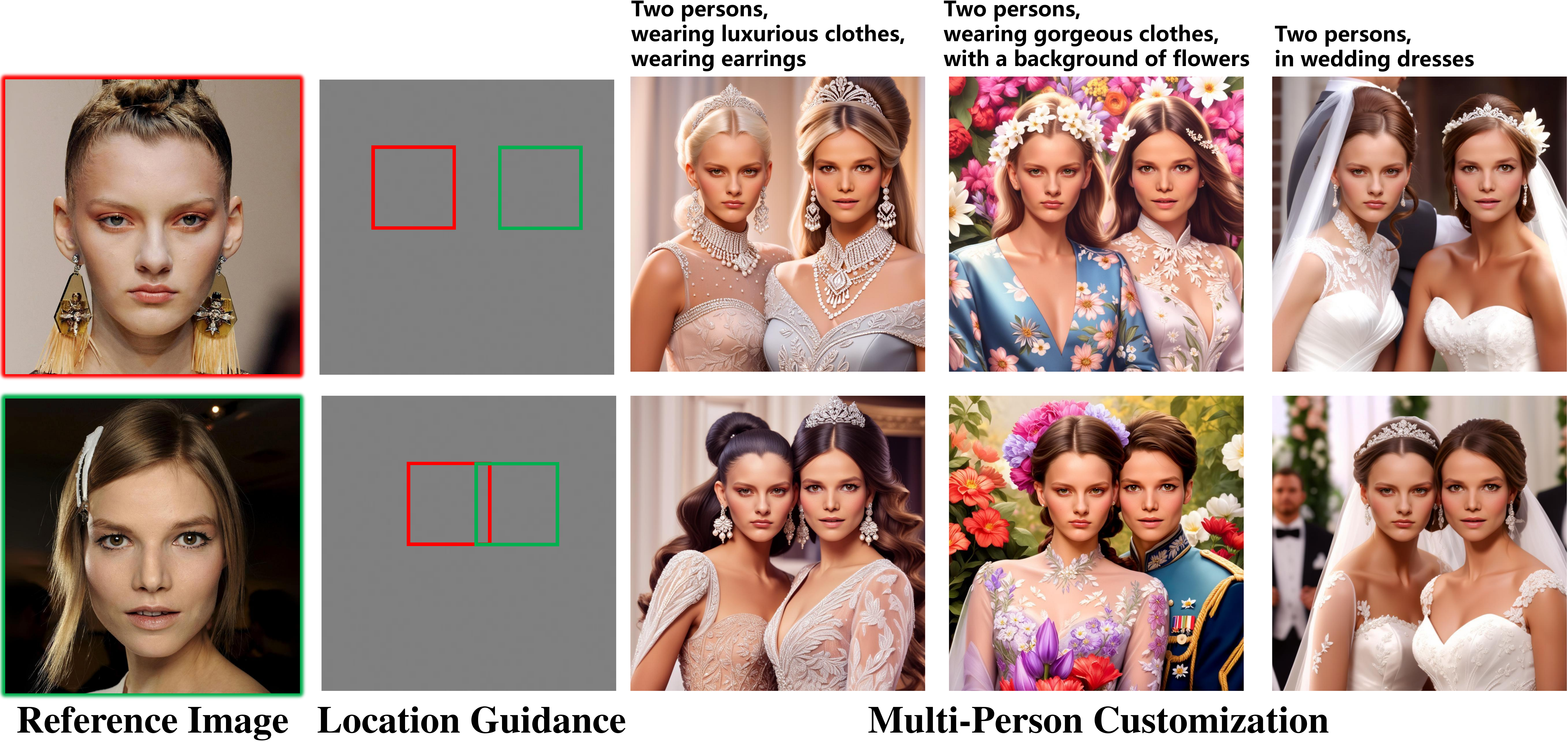}
	  \caption{Results of multi-person customization.
	  }
	  \label{fig:multi_person}
	\end{figure}

\subsubsection{Scale Robustness}
	To study the scale robustness of our RealisID, we vary the relative size of target faces from 1/7 to 1/4, and compare the generation performance of InstantID and our method. As depicted in Fig.~\ref{fig:scale}, compared to InstandID, our method is more effective in preserving facial structure and details across different face sizes. Table \ref{table:scale} further confirms that RealisID consistently surpasses InstantID in identity fidelity (measured by FaceNet metric) across all sizes, demonstrating its superior scale robustness.
	
\subsubsection{Fine Control}
	Our RealisID framework also provides fine control over other essential factors of the target face. Specifically, the location and size can be manipulated by manually setting the face bounding box. The pose and expression can also be controlled using the facial landmarks of other images with the desired pose and expression. The corresponding generation results are shown in Fig.~\ref{fig:pose_exp}.

\subsubsection{Multi-Person Customization}
	Given multiple reference images and their corresponding location guidance, RealisID can realize multi-person customization by separately integrating the injection information of different inputs in our local and global branches, as shown in Fig.~\ref{fig:method}(c). In this subsection, we study the multi-person customization in both non-overlapping and overlapping situations. From the results displayed in Fig.~\ref{fig:multi_person}, it can be seen that our method maintains the identity independence of two persons, even when there is overlap in their face locations.

\section{Conclusion}
	In this paper, we propose RealisID, a scale-robust and fine-controllable method for identity customization. It introduces a newly designed local branch to achieve the fine control of facial details and the identity fidelity across different face sizes. In addition, through another global branch, the overall harmony of the entire output image can be ensured, meanwhile the face location and the corresponding body and background layouts can be appropriately manipulated. Extensive experiments have verified the effectiveness of each individual branch, and the superiority of our whole RealisID framework over existing state-of-the-art methods. Please refer to the supplementary materials for more results.

\section*{Acknowledgments}
This work was in part supported by the National Key Research and Development Program of China (Grant No. 2022ZD0160604), the National Natural Science Foundation of China (Grant No. 62176194), the Young Scientists Fund of the National Natural Science Foundation of China (Grant No. 62306219), the Key Research and Development Program of Hubei Province (Grant No. 2023BAB083), the Project of Sanya Yazhou Bay Science and Technology City (Grant No. SCKJ- JYRC-2022-76, SKJC-2022-PTDX-031), the Project of Sanya Science and Education Innovation Park of Wuhan University of Technology (Grant No. 2021KF0031), Alibaba Group through Alibaba Research Intern Program.

\bibliography{aaai25}
\clearpage

\appendix
\section{Details of Quantitative Comparison}
	To ensure the reproducibility of experimental results in the main text, we will provide more implementation details of our quantitative comparison in this section. 
	
	\begin{table}[ht]\small
		\resizebox{1.0\linewidth}{!}{
			\begin{tabular}{|r |r |r |r |r|}
				422.jpg & 1598.jpg & 1971.jpg &2563.jpg &2771.jpg  \\
				3920.jpg & 5206.jpg & 6162.jpg &6713.jpg &7200.jpg  \\
				8234.jpg & 9092.jpg & 9333.jpg &10094.jpg &10429.jpg  \\
				10704.jpg & 12282.jpg & 12597.jpg &12936.jpg &13472.jpg  \\
				14209.jpg & 14407.jpg & 14421.jpg &15943.jpg &16146.jpg  \\
				17639.jpg & 20239.jpg & 22772.jpg &23165.jpg &23634.jpg  \\
				24414.jpg & 24521.jpg & 25419.jpg &25862.jpg &26116.jpg  \\
				26909.jpg & 27432.jpg & 28178.jpg &28351.jpg &29325.jpg  \\
			\end{tabular}
		}
		\caption{Identity list. The filenames of 40 identities that are randomly selected for our quantitative comparison.}
		\label{table:IDs}
	\end{table}
	
	\noindent \textbf{Identity List.} 
	In our experiment, 40 identities (unseen during the training stage) from another CelebA-HQ \cite{CelebA_HQ} dataset are randomly selected for quantitative comparison. Their corresponding filenames in CelebA-HQ are listed in Table~\ref{table:IDs}.
	
	\noindent \textbf{Prompt List.} For a comprehensive evaluation, we prepare 35 text prompts that covering a variety of clothing, attributes, actions, backgrounds, and styles. We list all the prompts used in our quantitative comparison in Table\ref{table:prompts}.
	
	\noindent \textbf{Number of Synthesized Images.} During the evaluation, every competing method generates 2 images for each prompt of each identity listed above, resulting in a total of 2800 ($35\times40\times2=2800$) synthesized images per method for the calculation of the four metrics.

	\begin{table}[t]
		\resizebox{1.0\linewidth}{!}{
			\begin{tabular}{l|l}
				\toprule
				Category & Prompt  \\
				\midrule
				\multirow{7}{*}{Clothing} & A person wearing a Superman costume\\
				& A person wearing a spacesuit\\
				& A person wearing a red sweater\\
				& A person wearing a black suit\\
				& A person wearing a wedding dress\\
				& A person wearing a white T-shirt\\
				& A person wearing a red hat\\
				\midrule
				\multirow{7}{*}{Attribute} & A person wearing headphones\\
				& A person with short red hair\\
				& A person with long yellow hair and pink lipstick\\
				& A person with clown makeup\\
				& A person wearing earrings\\
				& A person wearing glasses\\
				& A person with heavy makeup\\
				\midrule
				\multirow{7}{*}{Action} & A person playing the cello\\
				& A person playing the guitar\\
				& A person dancing\\
				& A person playing basketball\\
				& A person riding a bicycle\\
				& A person giving a speech on stage\\
				& A person singing\\
				\midrule
				\multirow{7}{*}{Background} & A person outdoors with a grassy background\\
				& A person outdoors with a background full of flowers\\
				& A person in space wearing a spacesuit\\
				& A person outdoors in snowy weather\\
				& A person outdoors with a pyramid in the background\\
				& A person outdoors with a snowy mountain in the background\\
				& A person outdoors with a beach in the background\\
				\midrule
				\multirow{7}{*}{Style} & A person in oil painting style\\
				& A person in watercolor style\\
				& A person in ink wash painting style\\
				& A person in cyberpunk style\\
				& A person in anime style\\
				& A person in Greek sculpture style\\
				& A person in colorful graffiti style\\
				\bottomrule
			\end{tabular}
		}
		\caption{Prompt list. We list all the text prompts used in our quantitative comparison.}
		\label{table:prompts}
	\end{table}

	\begin{table*}[t]
		\centering
		\resizebox{1.0\linewidth}{!}{
			\begin{tabular}{l|l|c|c|c|c}
				\toprule
				& Base Model & Scale-Robust  &Location \& Size  & Pose \& Expression & Multi-Person Customization \\
				\midrule
				IP-Adapter-face-plus & SD-1.5 &  & & &   \\
				FlashFace & SD-1.5 &  & $\surd$ & &   \\
				PhotoMaker & SDXL-1.0 & & & &   \\
				InstantID & SDXL-1.0 & & $\surd$ & $\surd$ &   \\
				PuLID & SDXL-1.0 & & & &   \\
				RealisID (Ours) & SDXL-1.0 & $\surd$ &  $\surd$ & $\surd$ &  $\surd$ \\
				\bottomrule
			\end{tabular}
		}
		\caption{Properties of state-of-the-art ID customization methods. 
			"Scale-Robust": In both regular and small face situations, the FaceNet metric (measuring identity authenticity) exceeds 0.7.
			"Location \& Size": Support control of facial location and size.
			"Pose \& Expression": Support control of facial poses and expressions.
			"Multi-Person Customization":  Support multi-person customization.}
		\label{table:properties}
	\end{table*}

	\section{Properties of Competing Methods}
	Table \ref{table:properties} presents properties of five state-of-the-art ID customization baselines and our RealisID framework. From the table, only our method can achieve the scale robustness and realize the multi-person customization, demonstrating the flexibility and practicality of our method. It is worth noting that our method is only trained on a single-person dataset but can be easily extended to handle multi-person customization by integrating multi-source injection information.

	\begin{table}[t]
		\centering
		\resizebox{1.0\linewidth}{!}{
			\begin{tabular}{l|c|c|c|c}
				\toprule
				{Pose} & 1/4   & 1/5  & 1/6 & 1/7 \\
				\midrule
				InstantID &0.0313 &0.0316 & 0.0333&0.0339   \\
		RealisID  &\textbf{0.0187} & \textbf{0.0194}&\textbf{0.0205}& \textbf{0.0231} \\
				\bottomrule
			\end{tabular}
		}
		\caption{Quantitative comparison of pose control.}
		\label{table:pose_control}
	\end{table}

	\begin{table}[t]
		\centering
		\resizebox{1.0\linewidth}{!}{
			\begin{tabular}{l|c|c|c|c}
				\toprule
				{Expression} & 1/4   & 1/5  & 1/6 & 1/7 \\
				\midrule
				InstantID &0.2591 &0.2622 & 0.2704&0.2738   \\
		RealisID &\textbf{0.1792} & \textbf{0.1821}&\textbf{0.1966}& \textbf{0.2032} \\
				\bottomrule
			\end{tabular}
		}
		\caption{Quantitative comparison of expression control.}
		\label{table:exp_control}
	\end{table}

	\begin{table}[t]
		\centering
		\resizebox{1.0\linewidth}{!}{
			\begin{tabular}{l|c|c|c|c}
				\toprule
				{FaceNet} & 1/4   & 1/5  & 1/6 & 1/7 \\
				\midrule
				RealisID (single) &0.791 &0.787 &0.772&0.748   \\
		RealisID (multi) &0.788 &0.786 &0.772 &0.746  \\
				\bottomrule
			\end{tabular}
		}
		\caption{Quantitative comparison of multi-person customization.}
		\label{table:multi_person}
	\end{table}

	\section{The effectiveness on Fine Controls and Multi-person Customization}
	\noindent \textbf{Fine Controls.} For the pose and expression control, we randomly select another 40 face images from the CeleA-HQ dataset and extract their facial landmarks as pose and expression conditions. We then use the DECA\cite{DECA} method to estimate the pose and expression parameters for the generated images and the conditions, respectively. Finally, we calculate the L1-norm between the estimated parameters and achieve the following results (the lower the better). In Table \ref{table:pose_control} and Table \ref{table:exp_control}, we can find that our method consistently outperforms InstantID under different face scales.

	\noindent \textbf{Multi-person Customization.}
	In addition, we use the above mentioned 40 images and original selected 40 images to perform the multi-person customization. We use the FaceNet metric to measure the ID fidelity for those generated multi-person images. See Table \ref{table:multi_person}, we can find that our RealisID obtains similar FaceNet values under single-person and multi-person cases, showing it can achieve high ID fidelity for each single person for the multi-person customization.

	\section{More Results}
	
\begin{figure*}[t]
		\centering
		\includegraphics[width=1.0\linewidth]{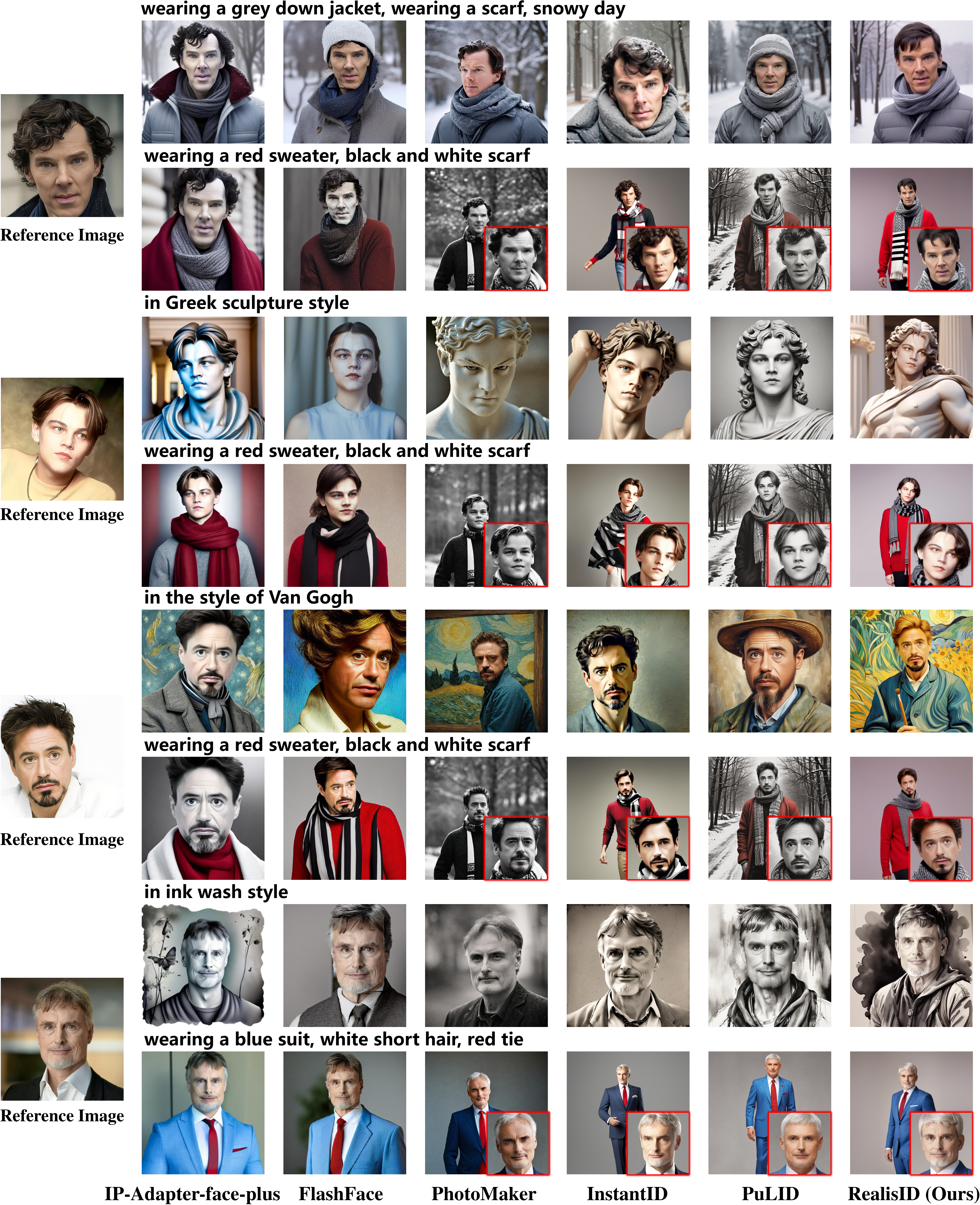}
		\caption{More qualitative comparison between different competing methods.
		}
		\label{fig:supp_qualitative_comparison}
	\end{figure*}

	\noindent \textbf{Qualitative Comparison.} More qualitative comparison results between different methods are shown in Fig.~\ref{fig:supp_qualitative_comparison}. It can be found that our RealisID framework generates comparable results with other state-of-the-art methods for the regular cases (see odd rows), and achieves high identity fidelity in the small face scenario (see even rows). In contrast, IP-Adapter-face-plus and FlashFace fail to produce small face images, while the other benchmark methods suffer some loss of identity information for the small face cases. This can further demonstrate the effectiveness of our method.

\begin{figure*}[t]
		\centering
		\includegraphics[width=1.0\linewidth]{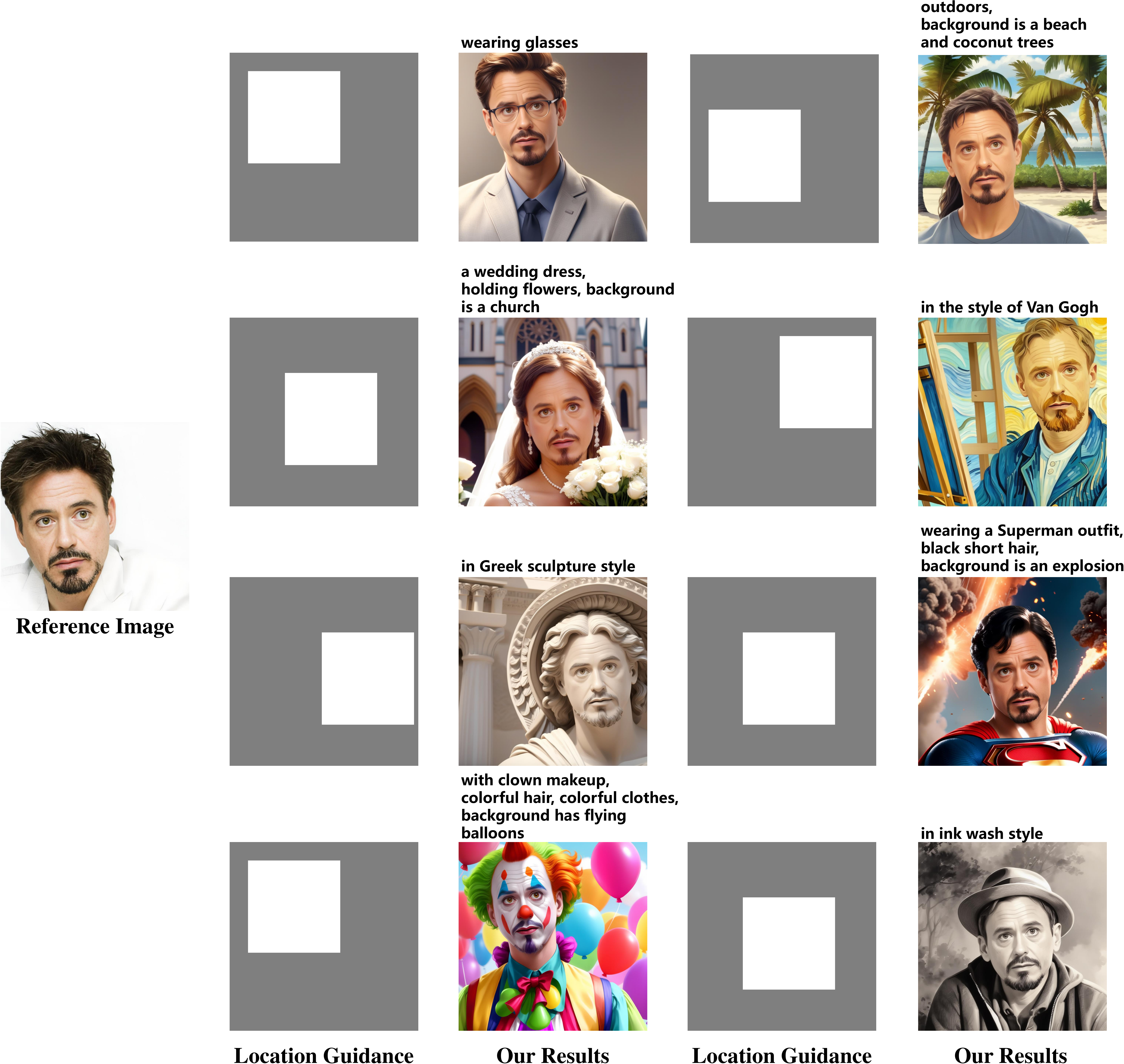}
		\caption{More results of our RealisID method for face location control.
		}
		\label{fig:supp_location}
	\end{figure*}

\begin{figure*}[t]
		\centering
		\includegraphics[width=1.0\linewidth]{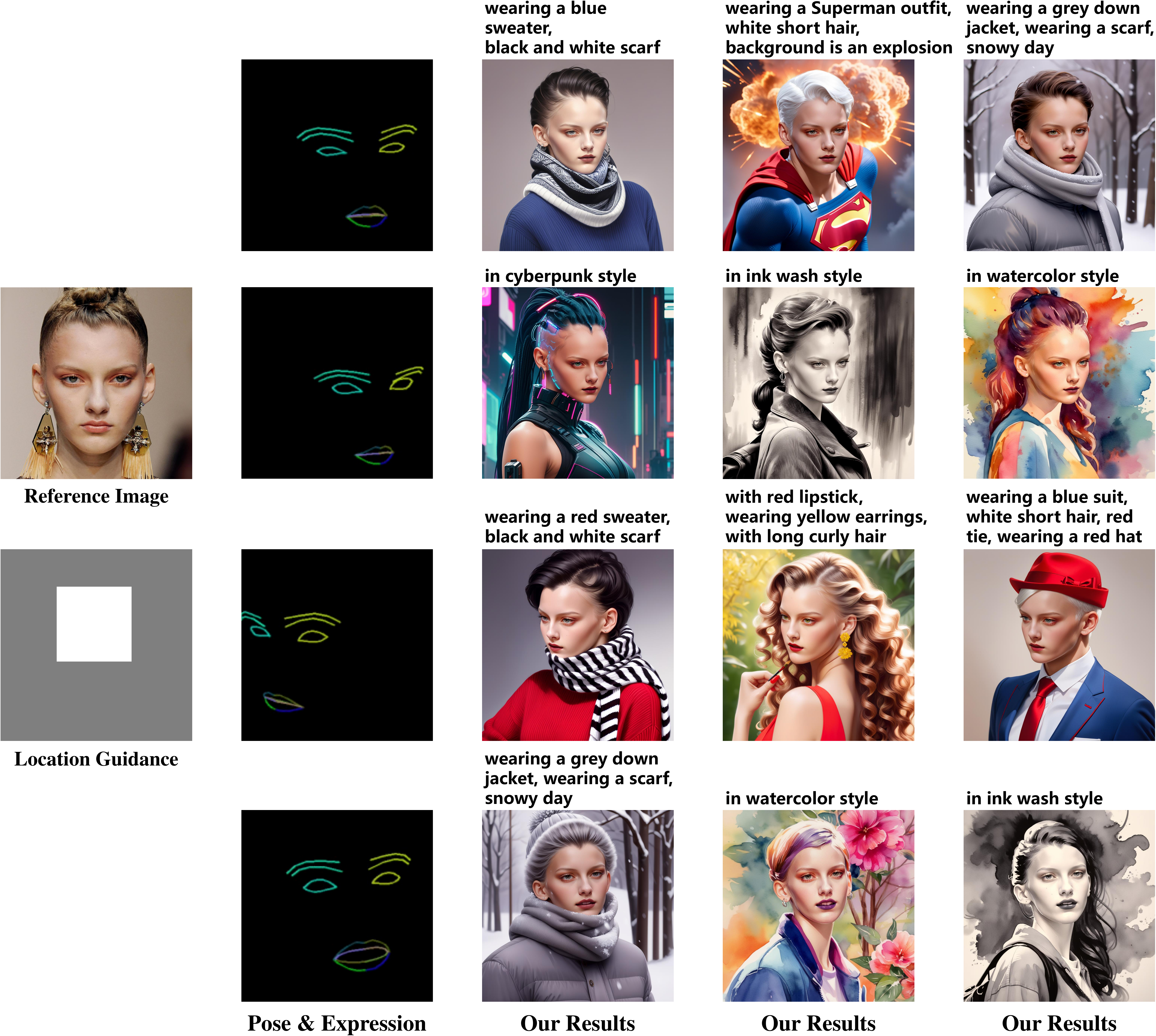}
		\caption{More results of our RealisID method for head pose control.
		}
		\label{fig:supp_pose1}
	\end{figure*}

\begin{figure*}[t]
		\centering
		\includegraphics[width=1.0\linewidth]{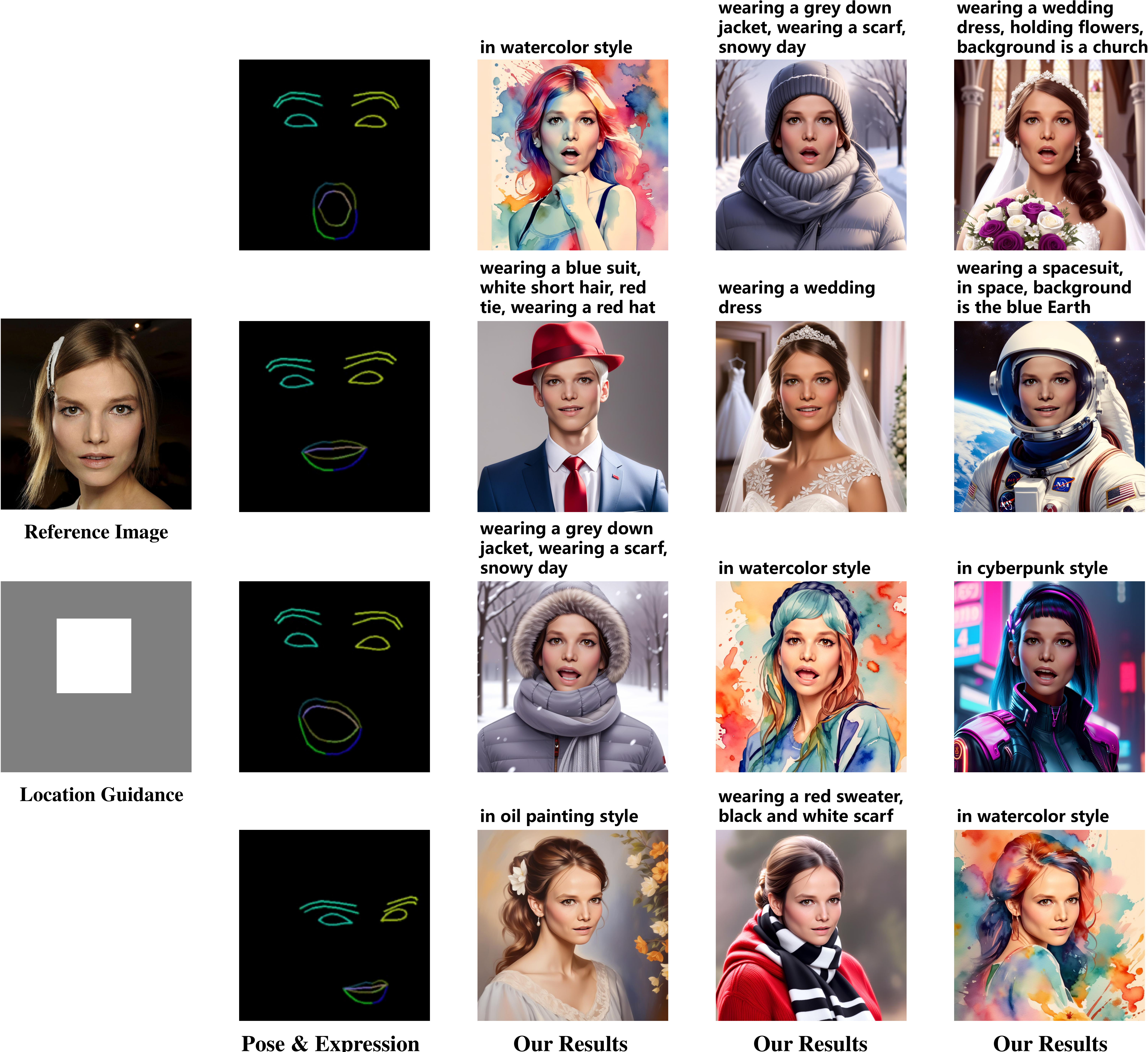}
		\caption{More results of our RealisID method for facial expression control.
		}
		\label{fig:supp_pose2}
	\end{figure*}

\begin{figure*}[t]
		\centering
		\includegraphics[width=1.0\linewidth]{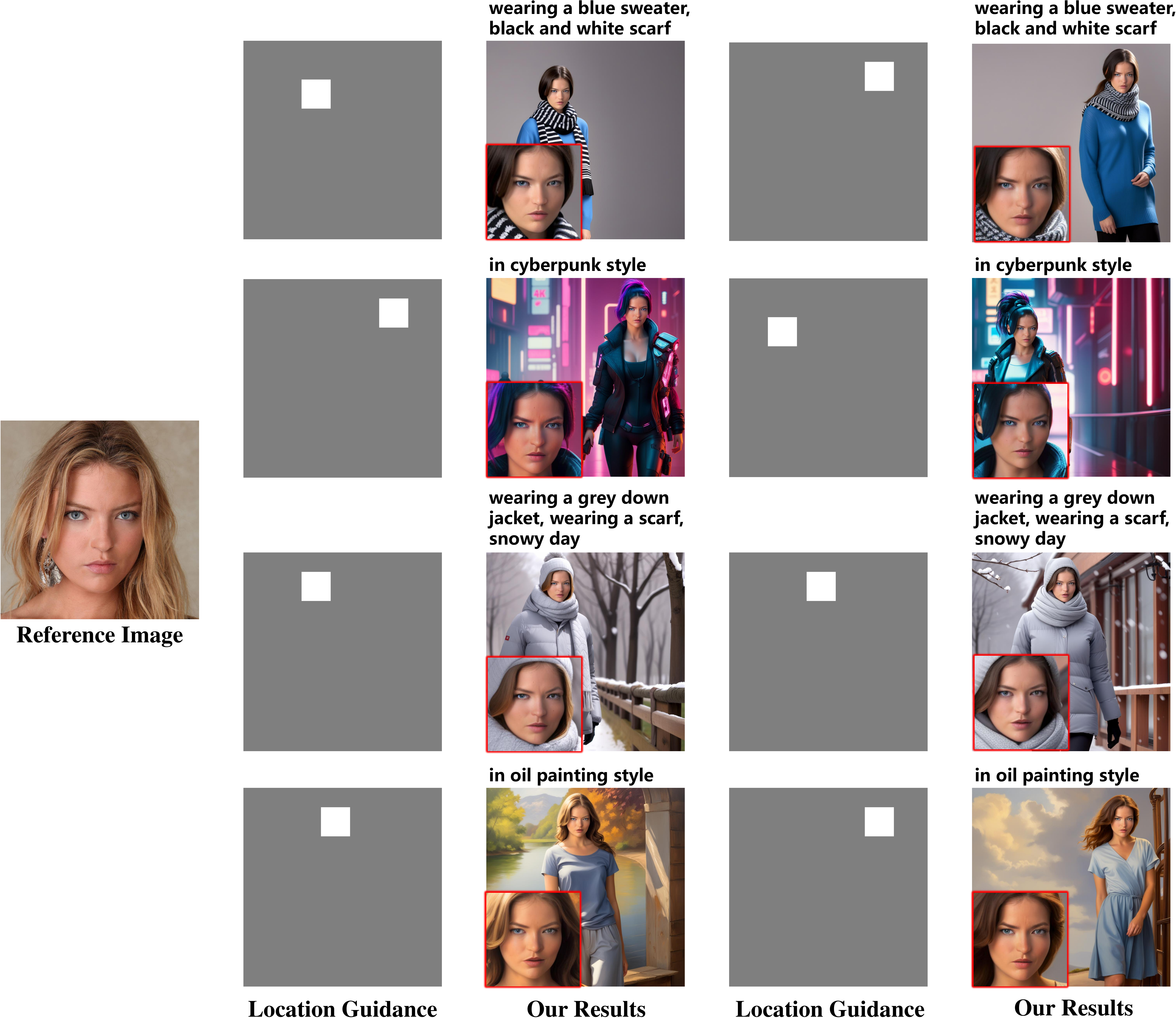}
		\caption{More results of our RealisID method on identity fidelity for small faces.
		}
		\label{fig:supp_small_face}
	\end{figure*}

	\noindent \textbf{Flexible and Fine Control.} We provide more results on the fine control of multiple facial factors, including face location (Fig.~\ref{fig:supp_location}), head pose (Fig.~\ref{fig:supp_pose1}), and facial expressions (Fig.~\ref{fig:supp_pose2}). We also present the small face images generated by our RealisID framework in Fig.~\ref{fig:supp_small_face}. All these results indicate that our method can generate high-quality customized portraits under flexible and fine controls.

\begin{figure*}[t]
		\centering
		\includegraphics[width=1.0\linewidth]{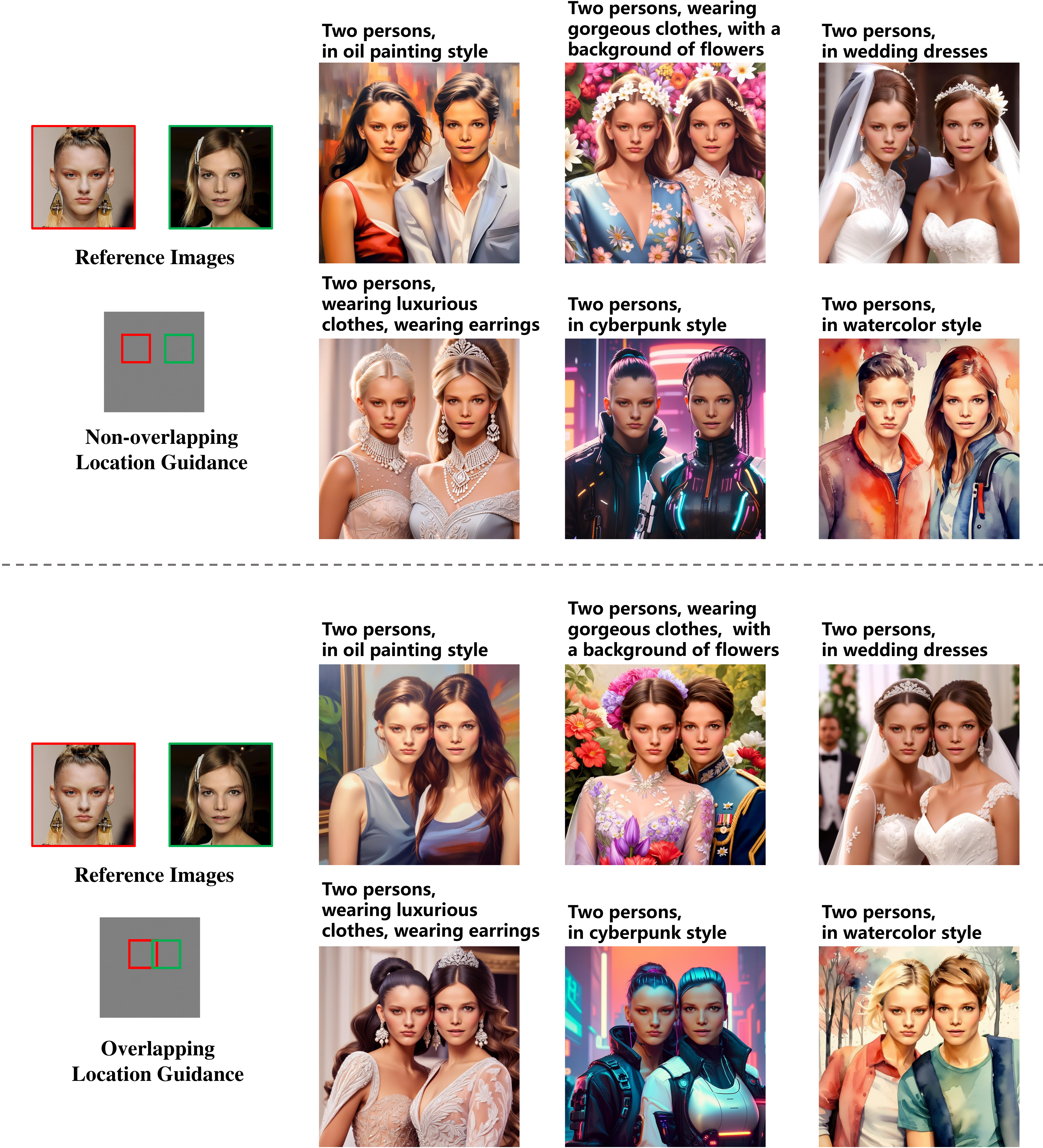}
		\caption{More results of our RealisID method for multi-person customization. 
		}
		\label{fig:supp_multi1}
	\end{figure*}

\begin{figure*}[t]
		\centering
		\includegraphics[width=1.0\linewidth]{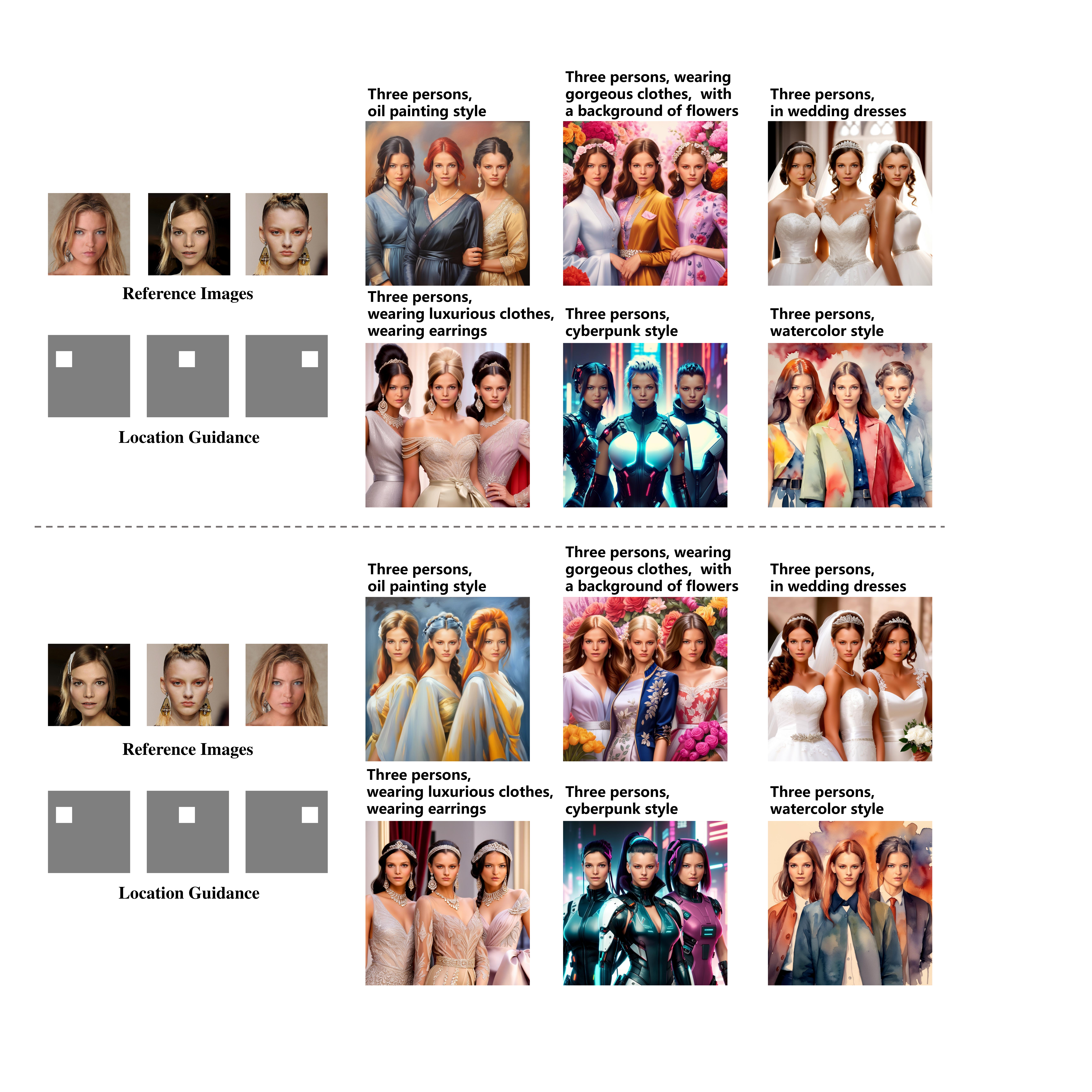}
		\caption{More results of our method on the face location control of each single individual for multi-person customization. 
		}
		\label{fig:supp_multi2}
	\end{figure*}

	\noindent \textbf{Multi-Person Customization.} More results of our method for multi-person customization are displayed in Fig.~\ref{fig:supp_multi1} and Fig.~\ref{fig:supp_multi2}. In Fig.~\ref{fig:supp_multi1}, we study the multi-person customization in both non-overlapping and overlapping situations. Our method maintains the identity independence of the two persons, even when there is overlap in their face locations. In Fig.~\ref{fig:supp_multi2}, we adjust the face location of each individual by changing the input order of their reference images.
	
\begin{figure*}[t]
		\centering
		\includegraphics[width=1.0\linewidth]{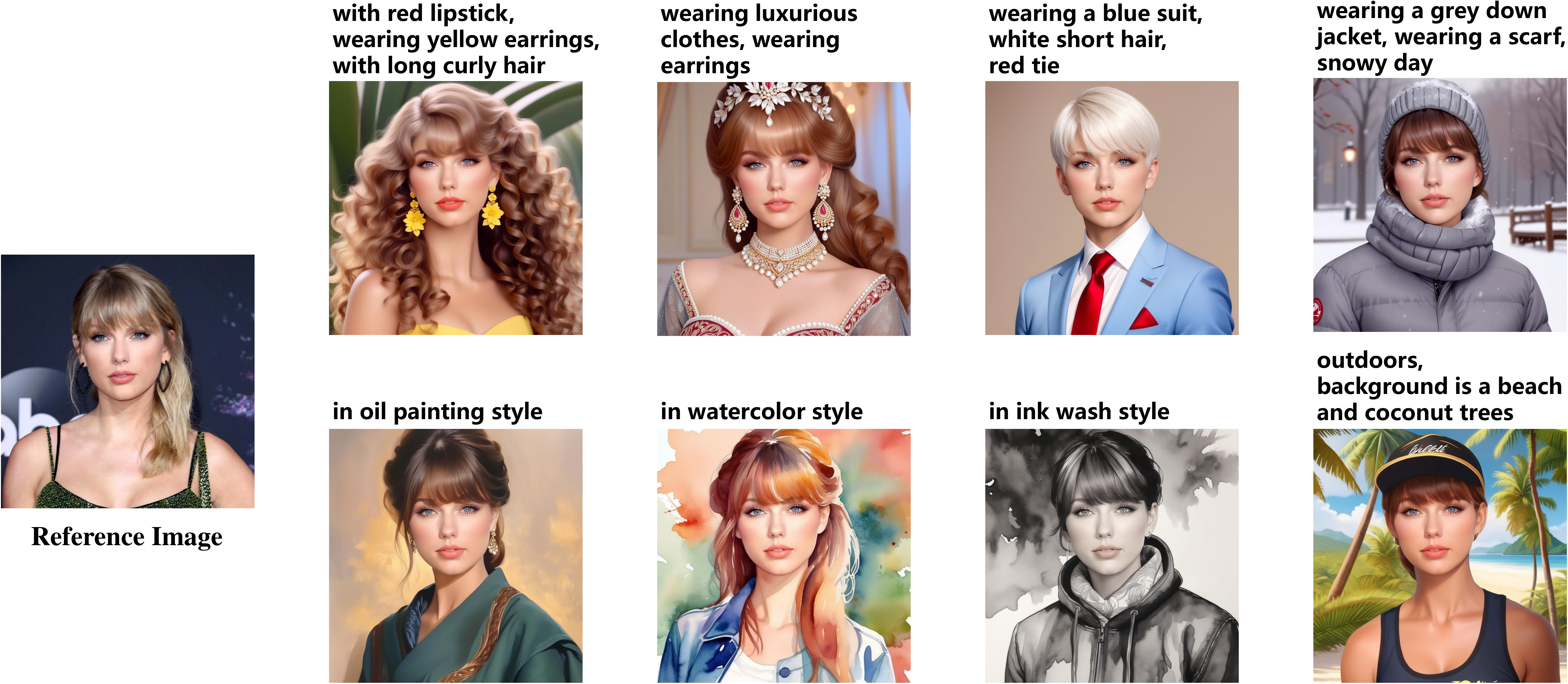}
		\caption{More results of our RealisID method under the guidance of diverse text prompts.
		}
		\label{fig:supp_prompt}
	\end{figure*}

	\noindent \textbf{Text Prompt Guidance.} By keep the pose-expression representation and location guidance same as the input reference image, our method can also work well with the guidance of different text prompts. The corresponding results are shown in Fig.~\ref{fig:supp_prompt}. It can be seen that diverse images that closely match the textual descriptions are generated.
	
	\begin{figure*}[t]
		\centering
		\includegraphics[width=1.0\linewidth]{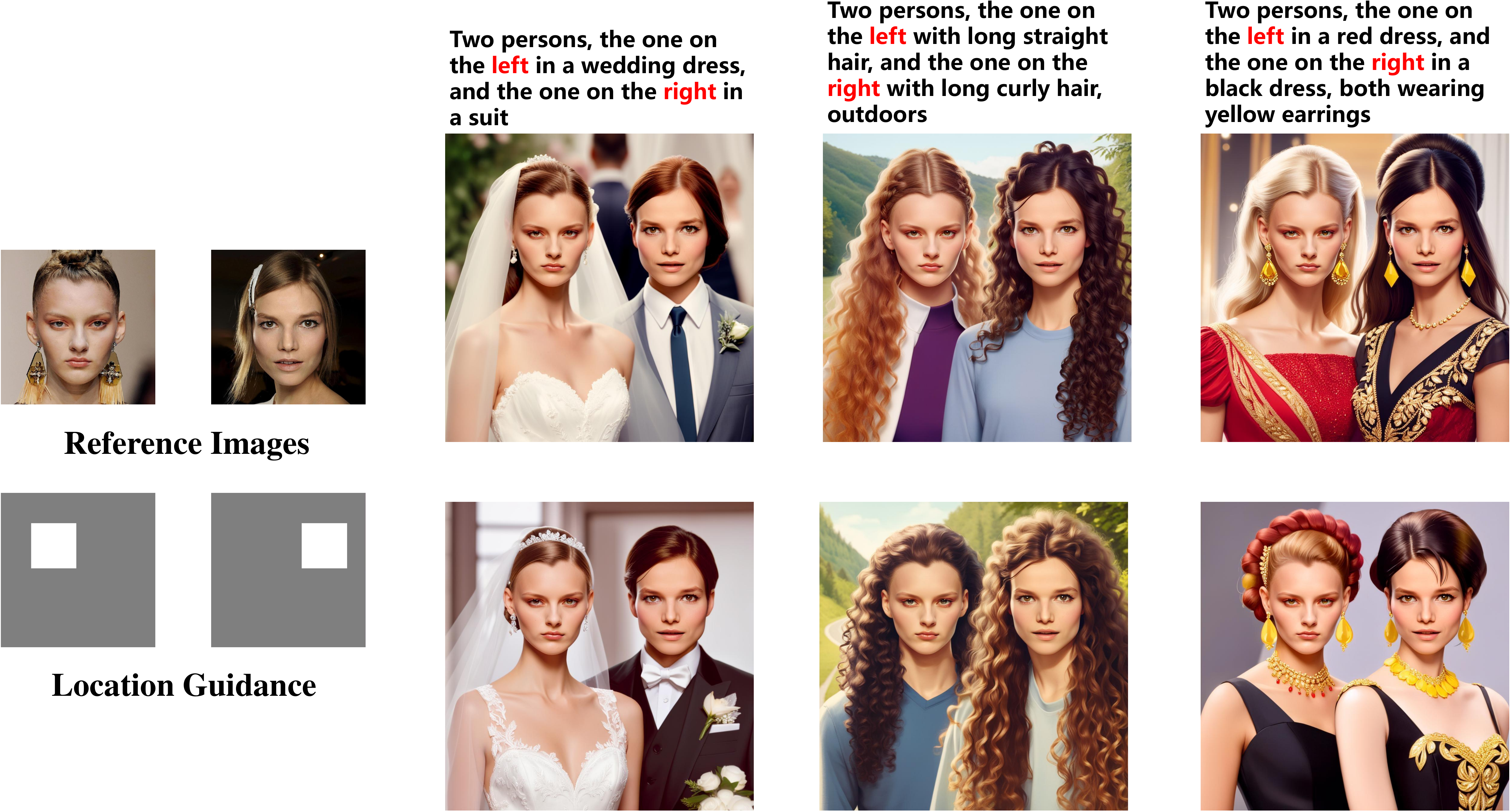}
		\caption{The limitation of our method. In multi-person customization, our method cannot achieve precise control over the face-irrelevant areas of each single person.
		}
		\label{fig:supp_limitation}
	\end{figure*}

	\section{Limitation}
		In multi-person customization, our method can achieve corresponding control over each person's facial region by separately modifying the pose-expression representation and location guidance in our local and global branches. But the generation of face-irrelevant information, such as image styles, is purely guided by the input text prompt, which is typically influence the entire image. Therefore, our method cannot precisely control the clothing, hairstyle, and other facial irrelevant areas for each single person. As shown in Fig.~\ref{fig:supp_limitation}, we attempt to control an individual's clothing and hairstyle by adding space related cues in the text prompts. However, due to the weak spatial understanding ability of existing text encoders, they have not been able to synthesize ideal results.

\end{document}